\documentclass[sigconf, review=False]{acmart}

\usepackage{booktabs} % For formal tables

\setcopyright{none}
\acmDOI{10.475/123_4}

\acmISBN{123-4567-24-567/08/06}

\acmConference[MLMH 2018]{KDD 2018}{August 2018}{London, United Kingdom}
\acmYear{2018}
\copyrightyear{2018}
\usepackage{graphicx} % more modern
\usepackage{subfigure} 

\usepackage{a4wide}

\def\eg{{\em e.g.,~}}
\def\ie{{\em i.e.,~}}

\def\reals{{\mathbb R}}

\def\cE{{\mathcal E}}
\def\cS{{\mathcal S}}

\definecolor{darkred}{rgb}{.8,.0,.035}

\newcommand{\deq}{:=}

\newcommand{\T}{\top}
\newcommand{\wh}{\widehat}

\begin{document}
\title{Multi-Task Learning with Incomplete Data for Healthcare}
%\titlenote{Produces the permission block, and
%  copyright information}
%\subtitle{Extended Abstract}
%\subtitlenote{The full version of the author's guide is available as
%  \texttt{acmart.pdf} document}
            
\author{Xin J. Hunt}
%\orcid{1234-5678-9012}
\affiliation{%
  \institution{SAS Institute Inc.}
  \streetaddress{100 SAS Campus Dr.}
  \city{Cary}
  \state{North Carolina}
  \postcode{27513}
}
\email{xin.hunt@sas.com}

\author{Saba Emrani}
%\orcid{1234-5678-9012}
\affiliation{%
  \institution{SAS Institute Inc.}
  \streetaddress{100 SAS Campus Dr.}
  \city{Cary}
  \state{North Carolina}
  \postcode{27513}
}
\email{semrani@ncsu.edu}

\author{Ilknur Kaynar Kabul}
%\authornote{The secretary disavows any knowledge of this author's actions.}
\affiliation{%
  \institution{SAS Institute Inc.}
  \streetaddress{100 SAS Campus Dr.}
  \city{Cary}
  \state{North Carolina}
  \postcode{27513}
}
\email{ilknur.kaynarkabul@sas.com}

\author{Jorge Silva}
\affiliation{%
  \institution{SAS Institute Inc.}
  \streetaddress{100 SAS Campus Dr.}
  \city{Cary}
  \state{North Carolina}
  \postcode{27513}
}
\email{jorge.silva@sas.com}

% The default list of authors is too long for headers.
%\renewcommand{\shortauthors}{X. Hunt et al.}
\renewcommand{\shorttitle}{Multi-Task Learning for Incomplete Data}

\begin{abstract}
Multi-task learning is a type of transfer learning that trains multiple tasks simultaneously and leverages the shared information between related tasks to improve the generalization performance.
However, missing features in the input matrix is a much more difficult problem which needs to be carefully addressed. 
Removing records with missing values can significantly reduce the sample size, which is impractical for datasets with large percentage of missing values.
Popular imputation methods often distort the covariance structure of the data, which causes inaccurate inference. In this paper we propose using plug-in covariance matrix estimators to tackle the challenge of missing features. Specifically, we analyze the plug-in estimators under the framework of robust multi-task learning with LASSO and graph regularization, which captures the relatedness between tasks via graph regularization. We use the Alzheimer's disease progression dataset as an example to show how the proposed framework is effective for prediction and model estimation when missing data is present.
\end{abstract}

%
% The code below should be generated by the tool at
% http://dl.acm.org/ccs.cfm
% Please copy and paste the code instead of the example below.
%

%\begin{CCSXML}
%<ccs2012>
%<concept>
%<concept_id>10010147.10010257.10010258.10010262</concept_id>
%<concept_desc>Computing methodologies~Multi-task learning</concept_desc>
%<concept_significance>500</concept_significance>
%</concept>
%<concept>
%<concept_id>10010147.10010257.10010321</concept_id>
%<concept_desc>Computing methodologies~Machine learning algorithms</concept_desc>
%<concept_significance>500</concept_significance>
%</concept>
%<concept>
%<concept_id>10010147.10010257.10010258.10010262.10010277</concept_id>
%<concept_desc>Computing methodologies~Transfer learning</concept_desc>
%<concept_significance>300</concept_significance>
%</concept>
%<concept>
%<concept_id>10010147.10010257.10010293.10010307</concept_id>
%<concept_desc>Computing methodologies~Learning linear models</concept_desc>
%<concept_significance>300</concept_significance>
%</concept>
%</ccs2012>
%\end{CCSXML}
%
%\ccsdesc[500]{Computing methodologies~Multi-task learning}
%\ccsdesc[500]{Computing methodologies~Machine learning algorithms}
%\ccsdesc[300]{Computing methodologies~Transfer learning}
%\ccsdesc[300]{Computing methodologies~Learning linear models}
%


\maketitle

\section{Introduction}
\label{Introduction}
The conventional machine learning frameworks consider only one learner seeking
to solve a single task. In numerous applications, however, there are multiple tasks labeling the same data instances differently. When the tasks are related, the information learned from each task can be used to enhance the learning of other tasks. It is therefore beneficial to learn relevant tasks together simultaneously, as opposed to learning each task independently. 
Multi-task learning uses the intrinsic relations between multiple tasks to improve the generalization performance. It benefits all the tasks by leveraging the task relatedness and shared information across relevant tasks.

\subsection{Multi-task learning}
Extensive experimental results have shown advantages of multi-task
learning compared to single task in problems involving related tasks \cite{bakker2003task, caruana1998multitask, heskes2000empirical, thrun2012learning}. 
Multi-task learning has been effectively used in a wide variety of applications including object location and recognition in image processing \cite{Caruana:1997:ML:262868.262872}, speech classification \cite{parameswaran2010large}, data integration from different web directories \cite{quadrianto2010multitask}, identification of handwritten digits \cite{quadrianto2010multitask}, multiple microarray data integration in bioinformatics \cite{widmer2012multitask}. In medical applications, multi-task learning has been successfully used to predict disease progression and estimate personalized medical models \cite {zhou2011multi, xu2015formula, emrani2017prognosis, nie2017modeling}.

%\subsection{Problem formulation}
Assume we have a total number of $K$ different tasks. For each task $i=1,\cdots,K$, we have $n_i$ observations in the observation matrix $X_i \in\reals^{n_i\times p}.$ The labels for the observations is denoted by $y_i\in\reals^{n_i}.$ Let $W = [W_1,\cdots,W_K]\in\reals^{p\times K}$ be the model we want to learn such that 
$
y_i = X_i W_i + \epsilon_i.
$
Here $\epsilon_i$ is random noise, and we want to learn $W_i$ from the observations $X_i$ and outcome $y_i$.

The relationship among $K$ tasks is represented by an undirected graph, where each task is a node, and the two nodes are connected if the two tasks are related. To encode the connections, let $\cE$ be the set of edges in the graph. Define a matrix $R\in\reals^{K\times |\cE|}$, where , and
\begin{equation}
\begin{aligned}
R_{i,j} =  \begin{cases}
1, \mbox{ if  $i<k$, edge $j$ connects nodes $i, k$,}\\
-1, \mbox{ if  $i>k$, edge $j$ connects nodes $i, k$,}\\
0, \mbox{ otherwise.}
\end{cases}
\end{aligned}
\end{equation}

When no missing data is present, the model $W$ can be efficiently estimated by solving the following optimization problem
\begin{equation}
\begin{aligned}
\widetilde{W} = \arg\min_W \sum_{i=1}^K \frac{1}{2n_i} \|X_iW_i - y_i\|_2^2 +  \frac{\mu}{2} \|W\|_1 
+ \frac{\lambda}{2} \|WR\|_F^2
\end{aligned}
\label{eq:Wtilde}
\end{equation}

Note that there are other regression-based formulations for multi-task learning problems \cite{zhou2011malsar}. The difference lies in the various penalization terms used. %\todo{more details if needed.} 
The proposed framework using plug-in estimators for covariance estimates can be easily adapted to fit the other multi-task learning formulations.

\subsection{Model Alzheimer's disease progression with multi-task learning}
In this paper, we use the Alzheimer's disease data from ADNI as an example of how the plug-in covariance estimators can help improve the performance of multi-task learning for medical and healthcare data.
\footnote{Data used in the preparation of this experiment were obtained from the Alzheimer's Disease Neuroimaging Initiative (ADNI) database (\url{adni.loni.usc.edu}). The ADNI was launched in
2003 as a public-private partnership, led by Principal Investigator Michael W. Weiner,
MD. The primary goal of ADNI has been to test whether serial magnetic resonance imaging
(MRI), positron emission tomography (PET), other biological markers, and clinical and
neuropsychological assessment can be combined to measure the progression of mild
cognitive impairment (MCI) and early Alzheimer's disease (AD). For up-to-date information,
see \url{www.adni-info.org}.}
The data $X_i$ is derived from the 1.5T MRI data of a total of 756 patients. The MRI images are preprocessed by UCSF using FreeSurfer, resulting in a total of 328 features. 
The objective is to predict the Alzheimer's Disease Assessment Scale cognitive subscale (ADAS-Cog) from the patients' MRI features. The ADAS-Cog score is the most popular cognitive testing instrument for evaluating the symptoms of Alzheimer's disease\cite{chu2000reliability, kolibas2000adas}. It includes questions on 11 tasks related to the patient's cognitive abilities like memory, language, attention, and praxis. In general, the higher the ADAS-Cog score are, the more significant the symptoms are.

The MRI data ($X_i$) is collected at each patient's first visit (month number 0). The $y_i$ includes the patients' ADAS-Cog scores at five visits over a total of two years, each spaced six months apart. We treat each visit as a task, where tasks 1 through 5 are the patients ADAS-Cog score at month number 0, 6, 12, 18, and 24, respectively. %Table~\ref{tab:AZpMiss} shows the number of patients and percentage of missing entries in each of the tasks. 
The number of patients for each task varies, however the feature vector for each individual patient remains constant for all five tasks. %The dataset has about 9\% of missing entries.

\begin{figure}[h!]
	\begin{center}
		\centerline{\includegraphics[width=.6\columnwidth]{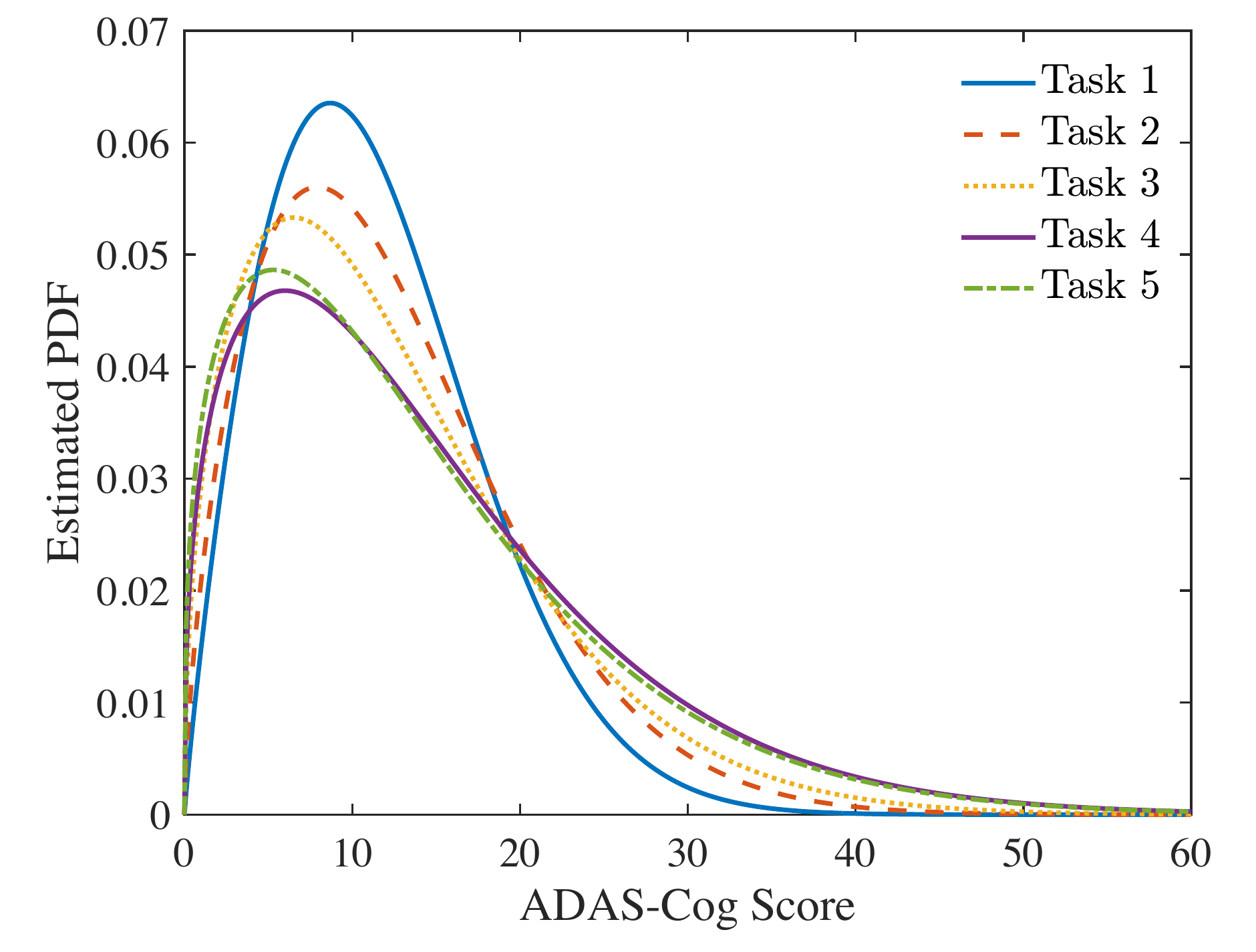}}
		\caption{Estimated Weibull distributions of ADAS-Cog scores of patients in each task of the Alzheimer's dataset. The gradual shifts of the estimated PDF reflects the progression of Alzheimer's disease in the patients.}
		\label{fig:weibullEst}
	\end{center}
	\vskip -0.2in
\end{figure}

\begin{figure}[t]
	\vskip 0.2in
	\begin{center}
		\centerline{\includegraphics[width=.7\columnwidth]{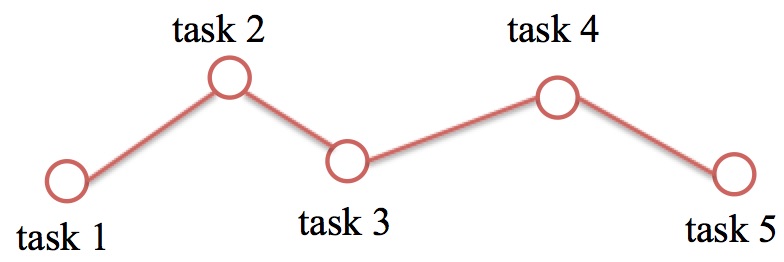}}
		\caption{The graph model of the five-task problem for Alzheimer's disease progression and experiments in Section~\ref{sec:SynExp}. Each task is related to its previous and subsequent tasks.}
		\label{fig:graphSyn}
	\end{center}
	\vskip -0.2in
\end{figure}

There are various ways to impose the relatedness of the tasks in order to incorporate the insight from task connections when learning all tasks jointly. For disease progression like this Alzheimer's disease case, due to the temporal relationship between the tasks, previous works \cite{zhou2011multi, emrani2017prognosis, nie2017modeling} have assumed a graph model where tasks are connected in temporal orders (as shown in Figure~\ref{fig:graphSyn}). However, is this intuitive model accurate?
To answer this question, we fit a Weibull distribution to the scores of each task ($y_i$), and plot the PDF of the five distributions in Figure~\ref{fig:weibullEst}. As seen in Figure~\ref{fig:weibullEst}, the score distribution slowly shifts over each visit, where tasks close in time have distributions close to each other\footnote{With the exception of Task 3 seems to be closer in distribution to Task 5 than Task 4. This is potentially caused by the sharp drop of patient number in Task 5, which can introduce bias in the data of Task 5. However, since the distributions of Task 4 and Task 5 are extremely close, we keep the graph model in Figure~\ref{fig:graphSyn} due to its interpretability.}. This suggests that the graph model in Figure~\ref{fig:graphSyn} is suitable. Notice the latter tasks tend to have higher probabilities for high score numbers. This reflects the progression of Alzheimer's disease in the patients.

\subsection{Handling Incomplete data}

Medical and healthcare data often come with incomplete data. There are plenty of cases in clinical studies where some patients do not answer certain questions or measurements of some biospecimens are partly lost at various stages \cite{ibrahim1990incomplete}. %In bioinformatics, experimentation errors, image corruptions and damages to the slides cause missing gene expression samples \cite{wang2006missing}. 
In the above mentioned Alzheimer's disease dataset, about 9\% of the data is missing. % In electronic healthcare records, a large percentage of data can come incomplete, and sometimes it can even be difficult to distinguish between missing data and negative results \cite{wells2013strategies}. 

The majority of machine learning approaches are however not developed for incomplete data,
%Ignoring or improper handling of missing data can deteriorate the performance of learning methods to a great extent.
%Dealing with missing data correctly therefore leads to many theoretical and computational challenges.
% existing solutions
%Since there is no explicit technique for tackling incomplete feature sets in learning process, 
and ad hoc methods such as case deletion and single imputation are typically used to complete the data and then proceed with the learning algorithm \cite{schafer2002missing}.
Case deletion approach simply removes the feature vectors with missing values. For carefully collected clinical data that has only a small percentage of missing values, case removal can be effective \cite{zhou2011multi, xu2015formula,emrani2017prognosis,nie2017modeling}.
However, excluding information from the data can introduce bias to the data, and is not efficient for datasets with limited observations.
%However, excluding information from the model is normally not efficient particularly in datasets with limited size. Moreover, the complete part of the data after removing missing features can be statistically nonrepresentative and may result in bias or considerable error. 
Moreover, even if the case deletion method performs well in training, missing values in the test set are still unavoidable. We cannot remove testing data completely for the reason that some parts of features are missing.

% A more efficient, broadly employed set of methods is single imputation that replaces the missing data points with reasonable estimations. 
Single imputation replaces missing data with estimations. The most common single imputation frameworks use mean imputation by averaging the observed data or alternatively the last observations carried forward in longitudinal studies or streaming. %Alternatively, conditional mean is used when an estimate for the distribution of missing features is available given the observed data \cite{schafer2002missing}.
%*
%
%Matrix completion methods are imputation techniques that are exploited for recovering large portions of missing data.
Low rank matrix completion \cite{candes2009exact} is another popular choice to impute missing values.% using nuclear norm minimization. 
Matrix completion methods are effective for recovering large portions of missing data.
%This algorithm performs under the assumption that the matrix to be recovered is low-rank or approximately low-rank and the missing values are distributed uniformly at random. %The low rank assumption however does not hold in many cases including medical records data. 
%Imputation methods are also not desirable since they neglect the uncertainty of missing values by replacing them with fixed instances, induce bias and underrate data variability \cite{cook2004marginal,jansen2006analyzing}. 
However, imputation methods neglect the uncertainty of missing values by replacing them with fixed instances, induce bias and underrate data variability \cite{cook2004marginal,jansen2006analyzing}.

It is important to note that in tackling the issue of incomplete data, the objective in this paper is making accurate
inference and estimation of the underlying models, not to retrieve missing values. Imputation of missing data does not always positively affect inference. For instance, replacing missing samples with mean of the observations
% might be an accurate estimation, but
can artificially inflate correlation between data, which is undesirable for inference.
Accordingly, incomplete data cannot be properly addressed separately from the modeling, learning and scoring procedures. We therefore focus on handling missing features within the learning process.

\subsection{Multi-task learning with missing data}

In this section we propose using plug-in covariance estimators when missing data exists in the observations.
Expand Eq.~\eqref{eq:Wtilde} and replace the empirical covariance matrices $X_i^{\T} X_i/n_i$ and $X_i^{\T} y_i/n_i$ with their estimators $\Gamma_i$ and $\gamma_i$, we have the following optimization problem

\begin{equation}
\begin{aligned}
\wh{W} =\arg\min_{W} \sum_{i=1}^K \left(\frac{1}{2n_i} W_i^{\T} \Gamma_i  W_i \right.
\left.- \frac{1}{n_i} W_i^{\T}\gamma_i\right) +  \frac{\mu}{2} \|W\|_1
+ \frac{\lambda}{2} \|WR\|_F^2,
\end{aligned}
\label{eq:What}
\end{equation}

This step of replacement is similar to \cite{loh2011high}, where the authors proposed using unbiased covariance estimators for the single-task LASSO problem (with no graph penalization). \cite{loh2011high} shows that such plug-in estimators results in robust model estimation for single-task LASSO when the observations are corrupted with missing data. We extend the idea to the multi-task scenario, and propose in Section~\ref{sec:estimators} a new plug-in estimators which show better performance at high levels of missing data.

%\subsection{Plug-In Covariance Estimation}
\section{Plug-In Covariance Estimations}
\label{sec:estimators}
In this section, we present three plug-in estimators for $\Gamma_i$ and $\gamma_i$. The performance of the three estimators is evaluated in Section~\ref{sec:exp}.%Equation (\eqref{eq:Gamma}) represents an estimation of the covariance matrix of the incomplete data based on Robust Lasso without any imputation. %Using experimental validation, we show that Robust Lasso provides a better estimation of the covariance matrix compared to other popular methods in state of the art including zero imputation, mean imputation and matrix factorization.

\subsection{R-LGR}
The R-LGR plug-in estimators are first proposed in \cite{loh2011high}, where the $\Gamma_i$ and $\gamma_i$ are estimated using unbiased estimators, where
\begin{equation} 
\begin{aligned}
\Gamma_i^{(0)} &= \arg\min_{\Gamma\in\cS_d} \|\frac{1}{n}X_i^{\T} X_i - \Gamma\|_2^2, \\
\gamma_i^{(0)} &= \arg\min_{\gamma\in\reals^d} \|\frac{1}{n}X_i^{\T} y_i - \gamma\|_2^2,
\end{aligned}
\end{equation}
where $\cS_d$ is the set of $d\times d$ positive-semidefinite symmetric matrices. The solutions are
\begin{equation} 
\begin{aligned}
\Gamma_i^{(0)} = \frac{1}{n_i}Z_i^{\T} Z_i - \frac{\rho_i}{n_i} \mbox{diag}(Z_i), \quad \gamma_i^{(0)} = \frac{1}{n_i}Z_i^{\T} y_i,
\end{aligned}
\label{eq:Gamma}
\end{equation}
where $Z_i = \frac{1}{1-\rho_i} X_i,$ and $\rho_i$ is the probability of data missing in $X_i$.

\subsection{R-LGR1}
R-LGR works well for small percentage of missing data. However, when large percentage of missing data is present (\eg over 20\%), $\Gamma_i^{(0)}$ tend to over-estimate the autocovariance matrix value, and cause large bias in the estimation. To compensate for the bias, we introduce an $\ell_1$ penalization in the autocovariance estimation problem, where
\begin{equation} 
\begin{aligned}
\Gamma_i^{(1)} &= \arg\min_{\Gamma\in\cS_d} \|\frac{1}{n_i}X_i^{\T} X_i - \Gamma\|_2^2 + \delta\|\Gamma\|_1.
\end{aligned}
\label{eq:Gamma1_0}
\end{equation}
%Here $\cS_d$ is the set of $d\times d$ positive-semidefinite symmetric matrices. 
This solution can be obtained by soft-thresholding the eigenvalues of $\Gamma_i^{(0)}$. Let 
$\Gamma_i^{(0)}  = Q_i^{(1)} D_i^{(1)} \left(Q_i^{(1)}\right)^{\T}$
be the eigendecomposition of $\Gamma_i^{(0)}$ from Eq.~\eqref{eq:Gamma}, and 
${d_i}^{(1)}\deq \mbox{diag}{D_i}^{(1)}$
be the diagonal elements of $D_i^{(1)}$, we can soft-threshold ${d_i}^{(1)}$ and compute 
$$
\Gamma_i^{(1)} = Q_i^{(1)} \ \mbox{T}_{\delta}(d_i^{(1)}) \ \left(Q_i^{(1)}\right)^{\T},
$$
for $i=1,\cdots,K,$
where the thresholding function $T(\cdot)$ is defined as
$$
\left(\mbox{T}_{\delta}(u)\right)_{j} = \begin{cases} u_{j}-\delta,& u_{j}\ge \delta \\ 
0,&|u_{j}|<\delta \\ \delta-u_{j},& u_{j}\le -\delta \end{cases}, \ j = 1,\cdots,p.
$$
$\gamma_i^{(1)}$ is the same as $\gamma_i^{(0)}$.
This matrix LASSO formulation has previously been used for high-dimensional covariance matrix estimation \cite{lounici2014high}. However, \cite{lounici2014high} does not analyze the estimator for prediction or model estimation, and no experimental validation is provided.

\subsection{Remarks}
${\Gamma}_i^{(1)}$ is a generalization of ${\Gamma}_i^{(0)}$, since ${\Gamma}_i^{(1)} = {\Gamma}_i^{(0)}$ when $\delta = 0$ in Eq.~\eqref{eq:Gamma1_0}. The matrix $\ell_1$ penalization term is introduced to reduce the intrinsic dimension of the covariance estimate, which is especially useful when $p$ is close to or larger than $n_i$. In this paper, we do not restrict $p\ge n_i$. However, as shown in Section~\ref{sec:exp}, the $\ell_1$ penalization offers better estimation stability when the proportion of missing data ($\rho_i$) is large. %However, the computation of ${\Gamma}_i^{(1)}$ is also more time-consuming that ${\Gamma}_i^{(0)}$, due to the extra steps of eigendecomposition and thresholding.

The computation of ${\Gamma}_i^{(0)}$ is the simpler and less computationally expensive than ${\Gamma}_i^{(1)}$, since ${\Gamma}_i^{(1)}$ computes the eigendecomposition of $p\times p$ matrices.
%, while ${\Gamma}_i^{(2)}$ computes the singular value decomposition of $n_i\times p$ matrices. 
However, since we only need to compute the eigendecomposition once for each data matrix $X_i$, the cost is usually not prohibitive. with the development and use of efficient distributed algorithms for eigendecomposition, the added computational cost would be manageable.

The $\ell_1$ penalization used for ${\Gamma}_i^{(1)}$ can drive the eigenvalues of the autocovariance matrix to zero, resulting in a sparse covariance matrix estimation. This is useful when the data is low-rank, or when we know the data correlation has low-dimensional structures. 

\section{Experiments}\label{sec:exp}
In this section, we show the results of the robust LASSO with graph regularization (R-LGR and R-LGR1) algorithms with both synthetic and real datasets.  We compare the proposed algorithm with two benchmark methods: 
(a) Mean-Impute: fill in missing entries with variable mean, and
(b) MF-LGR: fill in missing entries using matrix factorization \cite{keshavan2009matrix}.
%\item Oracle: fill in missing values with their true values, and then estimate $W$ with standard LGR as shown in Eq.~\eqref{eq:Wtilde}. This method is not affected by the percentage of missing data.
For both (a) and (b), after the imputation, we estimate $W$ with standard LASSO with graph penalization (LGR) as shown in Eq.~\eqref{eq:Wtilde}
The experimental results are obtained using the projected gradient descent algorithm \cite{calamai1987projected}.% Algorithm~\ref{alg:robustLGR}.

\subsection{Synthetic Experiments}
\label{sec:SynExp}

%In this experiment, we compare R-LGR, R-LGR1, R-LGR2, and the mean-impute estimator on their performance of prediction the model $W$. 
In the experiment, we evaluate the performance of the estimators on model and autocovariance estimation. Because there is no ground truth model or autocovariance matrix for real data with missing values, we perform these experiments with randomly generated data. Specifically, we generate five tasks with a graph structure illustrated in Figure~\ref{fig:graphSyn}. The ambient dimension is $p=50$, and we randomly generate $n_i=100$ observations for $i=1,\cdots,5$. For each task, $W_i$ is 7-sparse (\ie $W$ is 35-sparse). We randomly remove 5\% to 40\% of entries in the data. Each point in the plots is averaged over one hundred random realizations. All algorithms are tuned for minimum NMSE.

\begin{figure}[t]
	\vskip 0.2in
	\begin{center}
		\subfigure[Model estimation error]{\includegraphics[width=.5\columnwidth]{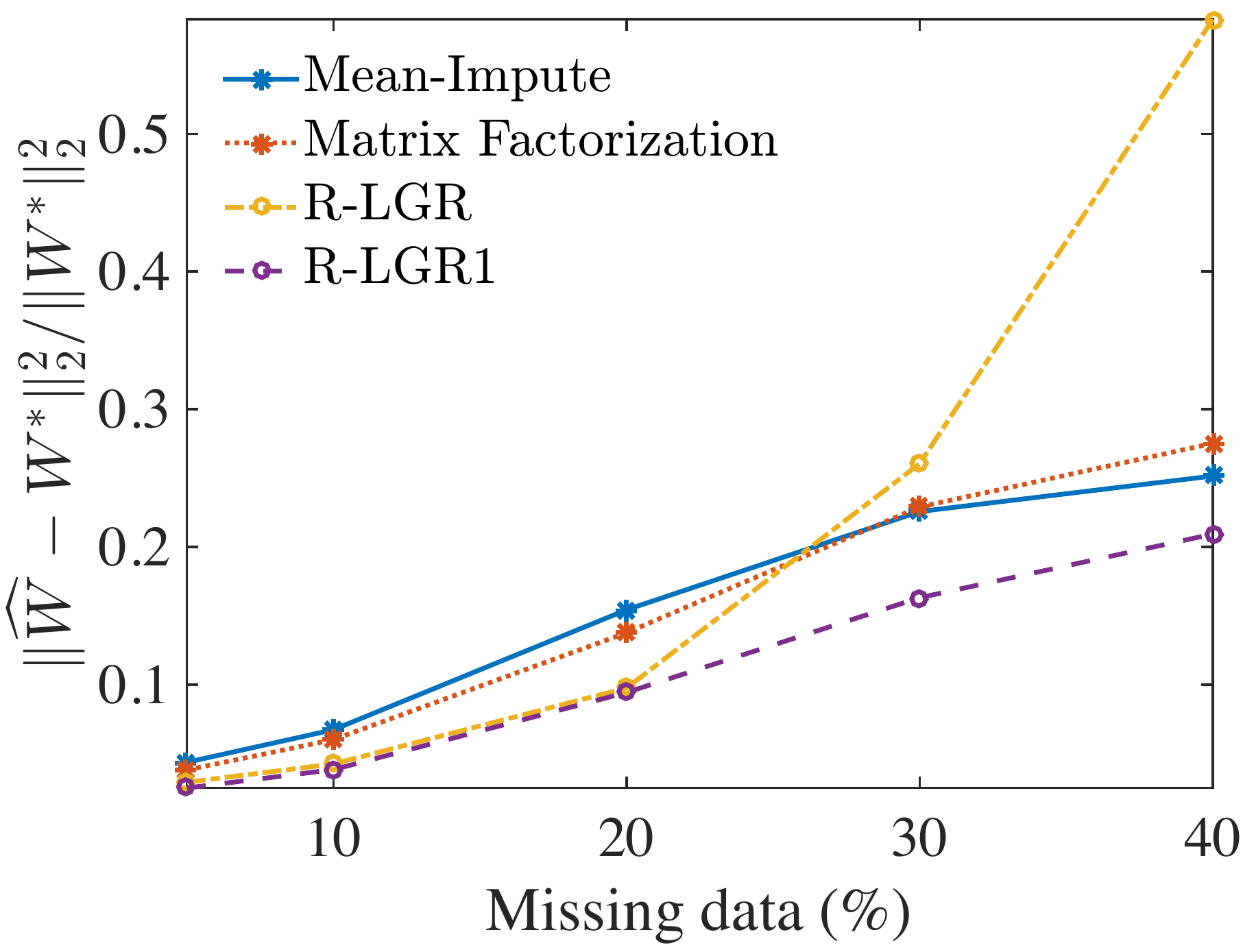}\label{fig:synExp1_1}}~
		\subfigure[Autocovariance matrix estimation error]{\includegraphics[width=.5\columnwidth]{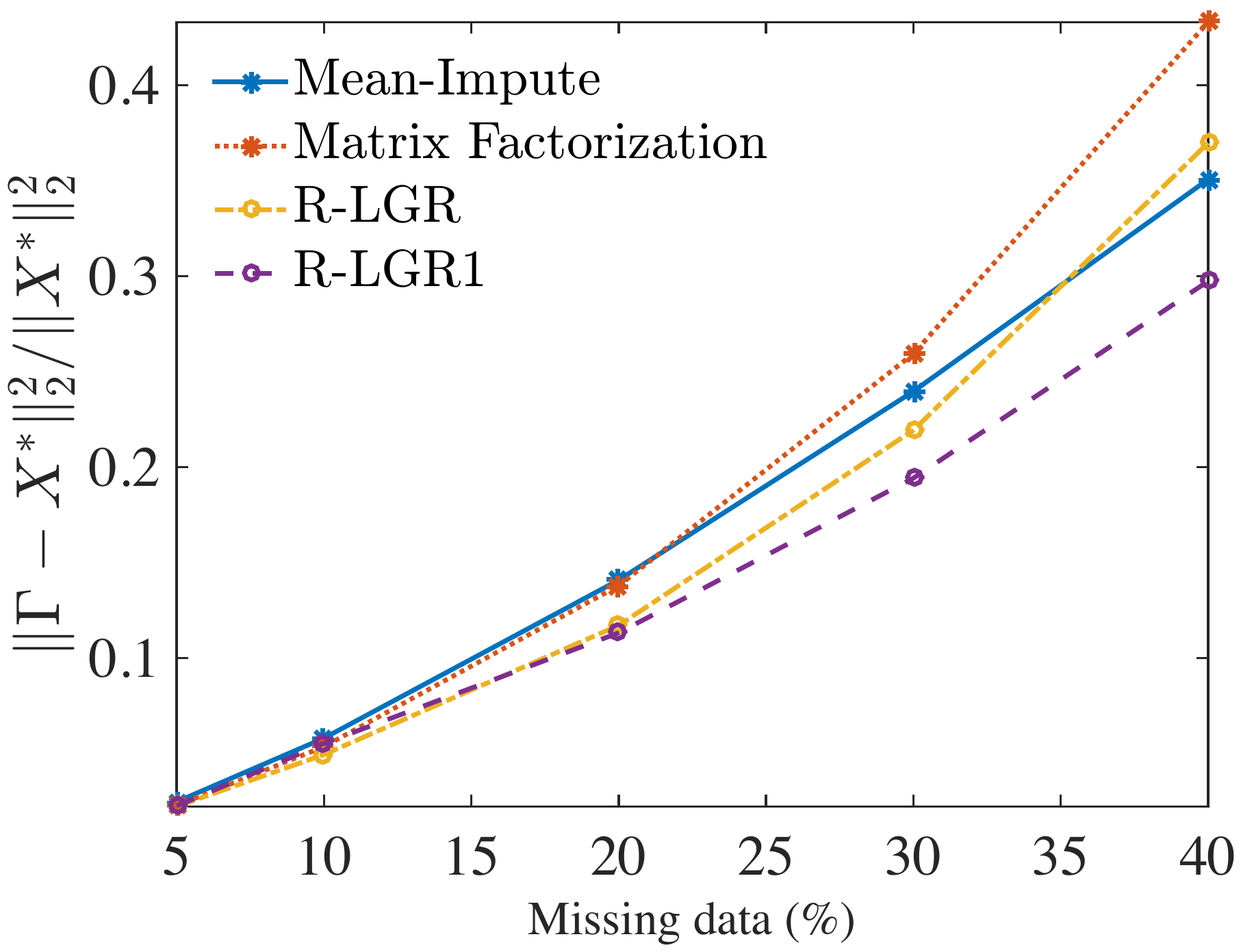}\label{fig:synExp1_2}}
		\caption{NMSE of $\wh{W}$ and $\Gamma$ as a function of the percentage of missing data.}
		\label{fig:synExp1}
	\end{center}
	\vskip -0.2in
\end{figure}

Figure~\ref{fig:synExp1_1} shows the NMSE of $\wh{W}$. Figure~\ref{fig:synExp1_2} shows the NMSE of the autocovariance estimator $\Gamma$. For mean-impute and matrix factorization, we compute $\Gamma = \widehat{X}^{\T}\widehat{X}$ after the imputation. In both experiments, R-LGR1 outperform mean-impute and matrix factorization methods for all missing data percentages. R-LGR works as well as R-LGR1 for small percentages of missing data. However, when the percentage of missing data exceeds 20\%, the R-LGR introduces more bias into the model estimator, and the NMSE for $\wh{W}$ grows larger than other competing algorithms, even though the error on $\Gamma$ of R-LGR stays similar to other methods.

\begin{table*}[h]
	\caption{Prediction error (RMSE) of ADAS-Cog score on the Alzheimer's dataset}
	\label{tab:predAlz}
	\vskip -0.1in
	\begin{center}
			%\begin{sc}
				\begin{tabular}{c|c|c|c|c|c}
					\hline
					%\abovespace\belowspace
					Missing Data & 9\% (Original Data) &19\% &29\% &39\% &49\% \\
					\hline
					%\abovespace\belowspace
					Mean-Impute & 7.596e-04 & 7.622e-04 & 7.661e-04 & 7.704e-04 & 7.737e-04 \\
					\hline
					MF-LGR &7.670e-04 & 7.712e-04 & 7.684e-04 & 7.692e-04 & 7.721e-04 \\
					%\belowspace 
					\hline
					%\abovespace\belowspace
					R-LGR &7.596e-04 & 7.622e-04 & 7.661e-04 & 7.704e-04 & 7.737e-04 \\
					%\belowspace 
					\hline
					%\abovespace\belowspace
					\bf{R-LGR1} & \bf{7.538e-04} & \bf{7.535e-04} & \bf{7.586e-04} & \bf{7.656e-04} & \bf{7.660e-04} \\
%					R-LGR1 & $5.033\times 10^{-4}$ \\
					\hline
					%\abovespace\belowspace
					%R-LGR2 & 7.565e-04 & 9.036e-04 & 1.377e-03 & 2.048e-03 & 2.061e-03  \\
%					R-LGR2 & $\st{5.027\times 10^{-4}}$ \\
					%\hline
				\end{tabular}
			%\end{sc}
	\end{center}
	\vskip -0.1in
\end{table*}

\subsection{Alzheimer's Disease Experiment}
\label{sec:AZexp}
In this experiment, we evaluate the prediction accuracy of the proposed estimators on the Alzheimer's disease dataset. 
We randomly divide the data into 60\% of training and 40\% of testing sets. The original dataset has about 9\% of missing entries. To access the plug-in estimators' stability with higher percentage of missing data, we also run experiments with randomly added 10\% to 40\% extra missing entries. We normalize all the features by $\ell_2$ norm. The prediction NMSE (averaged over twenty random realizations) is summarized in Table~\ref{tab:predAlz}. The first column shows results on the original data, while the second through the fifth column show results with added missing entries. All algorithms are tuned for minimum prediction NMSE ($\sum_{i=1}^K\frac{\|\hat{y_i} - y^*_i\|^2}{\|y^*_i\|^2}$). R-LGR1 outperforms all other competing methods for all levels of missing data tested.

\section{Conclusion}
\label{sec:conclusion}
In this study, a novel method for handling missing data with multi-task learning is proposed. Graph regularization is used as a general powerful tool to capture relatedness between connected tasks. To avoid bias and inaccurate inferences, the missing values are not handled separately from the modeling as performed in imputation and matrix completion approaches. We exploit plug-in estimators for the covariance matrices to handle missing features within the learning process. We also demonstrate that the proposed plug-in estimators provide better accuracy in model estimation and prediction accuracy when compared to other state-of-the-art imputation methods. Experimental results using synthetic and the Alzheimer's dataset show promising results for the R-LGR1 estimator.

In the future we intend to pursue research in two directions. First, we will expand the framework of plug-in covariance estimators for more multi-task learning methods with different objective function. Second, we will investigate data-driven methods to effectively generate the $R$ matrix for the graph model, especially in the presence of missing data.

\bibliographystyle{ACM-Reference-Format}
\bibliography{Multi-Missing}

%%% -*-BibTeX-*-
%%% Do NOT edit. File created by BibTeX with style
%%% ACM-Reference-Format-Journals [18-Jan-2012].

\begin{thebibliography}{24}

%%% ====================================================================
%%% NOTE TO THE USER: you can override these defaults by providing
%%% customized versions of any of these macros before the \bibliography
%%% command.  Each of them MUST provide its own final punctuation,
%%% except for \shownote{}, \showDOI{}, and \showURL{}.  The latter two
%%% do not use final punctuation, in order to avoid confusing it with
%%% the Web address.
%%%
%%% To suppress output of a particular field, define its macro to expand
%%% to an empty string, or better, \unskip, like this:
%%%
%%% \newcommand{\showDOI}[1]{\unskip}   % LaTeX syntax
%%%
%%% \def \showDOI #1{\unskip}           % plain TeX syntax
%%%
%%% ====================================================================

\ifx \showCODEN    \undefined \def \showCODEN     #1{\unskip}     \fi
\ifx \showDOI      \undefined \def \showDOI       #1{#1}\fi
\ifx \showISBNx    \undefined \def \showISBNx     #1{\unskip}     \fi
\ifx \showISBNxiii \undefined \def \showISBNxiii  #1{\unskip}     \fi
\ifx \showISSN     \undefined \def \showISSN      #1{\unskip}     \fi
\ifx \showLCCN     \undefined \def \showLCCN      #1{\unskip}     \fi
\ifx \shownote     \undefined \def \shownote      #1{#1}          \fi
\ifx \showarticletitle \undefined \def \showarticletitle #1{#1}   \fi
\ifx \showURL      \undefined \def \showURL       {\relax}        \fi
% The following commands are used for tagged output and should be
% invisible to TeX
\providecommand\bibfield[2]{#2}
\providecommand\bibinfo[2]{#2}
\providecommand\natexlab[1]{#1}
\providecommand\showeprint[2][]{arXiv:#2}

\bibitem[\protect\citeauthoryear{Bakker and Heskes}{Bakker and Heskes}{2003}]%
        {bakker2003task}
\bibfield{author}{\bibinfo{person}{B. Bakker} {and} \bibinfo{person}{T.
  Heskes}.} \bibinfo{year}{2003}\natexlab{}.
\newblock \showarticletitle{Task clustering and gating for bayesian multitask
  learning}.
\newblock \bibinfo{journal}{\emph{Journal of Machine Learning Research}}
  \bibinfo{volume}{4}, \bibinfo{number}{May} (\bibinfo{year}{2003}),
  \bibinfo{pages}{83--99}.
\newblock


\bibitem[\protect\citeauthoryear{Calamai and Mor{\'e}}{Calamai and
  Mor{\'e}}{1987}]%
        {calamai1987projected}
\bibfield{author}{\bibinfo{person}{P. Calamai} {and} \bibinfo{person}{J.
  Mor{\'e}}.} \bibinfo{year}{1987}\natexlab{}.
\newblock \showarticletitle{Projected gradient methods for linearly constrained
  problems}.
\newblock \bibinfo{journal}{\emph{Mathematical programming}}
  \bibinfo{volume}{39}, \bibinfo{number}{1} (\bibinfo{year}{1987}),
  \bibinfo{pages}{93--116}.
\newblock


\bibitem[\protect\citeauthoryear{Cand{\`e}s and Recht}{Cand{\`e}s and
  Recht}{2009}]%
        {candes2009exact}
\bibfield{author}{\bibinfo{person}{E.~J. Cand{\`e}s} {and} \bibinfo{person}{B.
  Recht}.} \bibinfo{year}{2009}\natexlab{}.
\newblock \showarticletitle{Exact matrix completion via convex optimization}.
\newblock \bibinfo{journal}{\emph{Foundations of Computational mathematics}}
  \bibinfo{volume}{9}, \bibinfo{number}{6} (\bibinfo{year}{2009}),
  \bibinfo{pages}{717--772}.
\newblock


\bibitem[\protect\citeauthoryear{Caruana}{Caruana}{1997}]%
        {Caruana:1997:ML:262868.262872}
\bibfield{author}{\bibinfo{person}{R. Caruana}.}
  \bibinfo{year}{1997}\natexlab{}.
\newblock \showarticletitle{Multitask Learning}.
\newblock \bibinfo{journal}{\emph{Mach. Learn.}} \bibinfo{volume}{28},
  \bibinfo{number}{1} (\bibinfo{date}{July} \bibinfo{year}{1997}),
  \bibinfo{pages}{41--75}.
\newblock
\showISSN{0885-6125}
\urldef\tempurl%
\url{https://doi.org/10.1023/A:1007379606734}
\showDOI{\tempurl}


\bibitem[\protect\citeauthoryear{Caruana}{Caruana}{1998}]%
        {caruana1998multitask}
\bibfield{author}{\bibinfo{person}{R. Caruana}.}
  \bibinfo{year}{1998}\natexlab{}.
\newblock \showarticletitle{Multitask learning}.
\newblock In \bibinfo{booktitle}{\emph{Learning to learn}}.
  \bibinfo{publisher}{Springer}, \bibinfo{pages}{95--133}.
\newblock


\bibitem[\protect\citeauthoryear{Chu, Chiu, Hui, Yu, Tsui, and Lee}{Chu
  et~al\mbox{.}}{2000}]%
        {chu2000reliability}
\bibfield{author}{\bibinfo{person}{L. Chu}, \bibinfo{person}{K. Chiu},
  \bibinfo{person}{S. Hui}, \bibinfo{person}{G. Yu}, \bibinfo{person}{W. Tsui},
  {and} \bibinfo{person}{P. Lee}.} \bibinfo{year}{2000}\natexlab{}.
\newblock \showarticletitle{The reliability and validity of the Alzheimer's
  Disease Assessment Scale Cognitive Subscale (ADAS-Cog) among the elderly
  Chinese in Hong Kong}.
\newblock \bibinfo{journal}{\emph{Annals of the Academy of Medicine,
  Singapore}} \bibinfo{volume}{29}, \bibinfo{number}{4} (\bibinfo{year}{2000}),
  \bibinfo{pages}{474--485}.
\newblock


\bibitem[\protect\citeauthoryear{Cook, Zeng, and Yi}{Cook
  et~al\mbox{.}}{2004}]%
        {cook2004marginal}
\bibfield{author}{\bibinfo{person}{R. Cook}, \bibinfo{person}{L. Zeng}, {and}
  \bibinfo{person}{G. Yi}.} \bibinfo{year}{2004}\natexlab{}.
\newblock \showarticletitle{Marginal analysis of incomplete longitudinal binary
  data: a cautionary note on LOCF imputation}.
\newblock \bibinfo{journal}{\emph{Biometrics}} \bibinfo{volume}{60},
  \bibinfo{number}{3} (\bibinfo{year}{2004}), \bibinfo{pages}{820--828}.
\newblock


\bibitem[\protect\citeauthoryear{Emrani, McGuirk, and Xiao}{Emrani
  et~al\mbox{.}}{2017}]%
        {emrani2017prognosis}
\bibfield{author}{\bibinfo{person}{S. Emrani}, \bibinfo{person}{A. McGuirk},
  {and} \bibinfo{person}{W. Xiao}.} \bibinfo{year}{2017}\natexlab{}.
\newblock \showarticletitle{Prognosis and Diagnosis of Parkinson's Disease
  Using Multi-Task Learning}. In \bibinfo{booktitle}{\emph{Proceedings of the
  23rd ACM SIGKDD International Conference on Knowledge Discovery and Data
  Mining}}. ACM, \bibinfo{pages}{1457--1466}.
\newblock


\bibitem[\protect\citeauthoryear{Heskes}{Heskes}{2000}]%
        {heskes2000empirical}
\bibfield{author}{\bibinfo{person}{T. Heskes}.}
  \bibinfo{year}{2000}\natexlab{}.
\newblock \showarticletitle{Empirical Bayes for learning to learn}. In
  \bibinfo{booktitle}{\emph{ICML}}. \bibinfo{pages}{367--374}.
\newblock


\bibitem[\protect\citeauthoryear{Ibrahim}{Ibrahim}{1990}]%
        {ibrahim1990incomplete}
\bibfield{author}{\bibinfo{person}{J. Ibrahim}.}
  \bibinfo{year}{1990}\natexlab{}.
\newblock \showarticletitle{Incomplete data in generalized linear models}.
\newblock \bibinfo{journal}{\emph{J. Amer. Statist. Assoc.}}
  \bibinfo{volume}{85}, \bibinfo{number}{411} (\bibinfo{year}{1990}),
  \bibinfo{pages}{765--769}.
\newblock


\bibitem[\protect\citeauthoryear{Jansen, Beunckens, Molenberghs, Verbeke,
  Mallinckrodt, et~al\mbox{.}}{Jansen et~al\mbox{.}}{2006}]%
        {jansen2006analyzing}
\bibfield{author}{\bibinfo{person}{I. Jansen}, \bibinfo{person}{C. Beunckens},
  \bibinfo{person}{G. Molenberghs}, \bibinfo{person}{G. Verbeke},
  \bibinfo{person}{C. Mallinckrodt}, {et~al\mbox{.}}}
  \bibinfo{year}{2006}\natexlab{}.
\newblock \showarticletitle{Analyzing incomplete discrete longitudinal clinical
  trial data}.
\newblock \bibinfo{journal}{\emph{Statist. Sci.}} \bibinfo{volume}{21},
  \bibinfo{number}{1} (\bibinfo{year}{2006}), \bibinfo{pages}{52--69}.
\newblock


\bibitem[\protect\citeauthoryear{Keshavan, Oh, and Montanari}{Keshavan
  et~al\mbox{.}}{2009}]%
        {keshavan2009matrix}
\bibfield{author}{\bibinfo{person}{R. Keshavan}, \bibinfo{person}{S. Oh}, {and}
  \bibinfo{person}{A. Montanari}.} \bibinfo{year}{2009}\natexlab{}.
\newblock \showarticletitle{Matrix completion from a few entries}. In
  \bibinfo{booktitle}{\emph{2009 IEEE International Symposium on Information
  Theory}}. IEEE, \bibinfo{pages}{324--328}.
\newblock


\bibitem[\protect\citeauthoryear{Kolibas, Korinkova, Novotny, Vajdickova, and
  Hunakova}{Kolibas et~al\mbox{.}}{2000}]%
        {kolibas2000adas}
\bibfield{author}{\bibinfo{person}{E. Kolibas}, \bibinfo{person}{V. Korinkova},
  \bibinfo{person}{V. Novotny}, \bibinfo{person}{K. Vajdickova}, {and}
  \bibinfo{person}{D. Hunakova}.} \bibinfo{year}{2000}\natexlab{}.
\newblock \showarticletitle{ADAS-Cog (Alzheimer's Disease Assessment
  Scale-cognitive subscale)-validation of the Slovak version}.
\newblock \bibinfo{journal}{\emph{Bratislavske lekarske listy}}
  \bibinfo{volume}{101}, \bibinfo{number}{11} (\bibinfo{year}{2000}),
  \bibinfo{pages}{598--602}.
\newblock


\bibitem[\protect\citeauthoryear{Loh and Wainwright}{Loh and
  Wainwright}{2011}]%
        {loh2011high}
\bibfield{author}{\bibinfo{person}{P.-L. Loh} {and} \bibinfo{person}{M.
  Wainwright}.} \bibinfo{year}{2011}\natexlab{}.
\newblock \showarticletitle{High-dimensional regression with noisy and missing
  data: Provable guarantees with non-convexity}. In
  \bibinfo{booktitle}{\emph{Advances in Neural Information Processing
  Systems}}. \bibinfo{pages}{2726--2734}.
\newblock


\bibitem[\protect\citeauthoryear{Lounici et~al\mbox{.}}{Lounici
  et~al\mbox{.}}{2014}]%
        {lounici2014high}
\bibfield{author}{\bibinfo{person}{K. Lounici} {et~al\mbox{.}}}
  \bibinfo{year}{2014}\natexlab{}.
\newblock \showarticletitle{High-dimensional covariance matrix estimation with
  missing observations}.
\newblock \bibinfo{journal}{\emph{Bernoulli}} \bibinfo{volume}{20},
  \bibinfo{number}{3} (\bibinfo{year}{2014}), \bibinfo{pages}{1029--1058}.
\newblock


\bibitem[\protect\citeauthoryear{Nie, Zhang, Meng, Song, Chang, and Li}{Nie
  et~al\mbox{.}}{2017}]%
        {nie2017modeling}
\bibfield{author}{\bibinfo{person}{L. Nie}, \bibinfo{person}{L. Zhang},
  \bibinfo{person}{L. Meng}, \bibinfo{person}{X. Song}, \bibinfo{person}{X.
  Chang}, {and} \bibinfo{person}{X. Li}.} \bibinfo{year}{2017}\natexlab{}.
\newblock \showarticletitle{Modeling disease progression via multisource
  multitask learners: A case study with Alzheimer's disease}.
\newblock \bibinfo{journal}{\emph{IEEE transactions on neural networks and
  learning systems}} \bibinfo{volume}{28}, \bibinfo{number}{7}
  (\bibinfo{year}{2017}), \bibinfo{pages}{1508--1519}.
\newblock


\bibitem[\protect\citeauthoryear{Parameswaran and Weinberger}{Parameswaran and
  Weinberger}{2010}]%
        {parameswaran2010large}
\bibfield{author}{\bibinfo{person}{S. Parameswaran} {and} \bibinfo{person}{K.
  Weinberger}.} \bibinfo{year}{2010}\natexlab{}.
\newblock \showarticletitle{Large margin multi-task metric learning}. In
  \bibinfo{booktitle}{\emph{Advances in neural information processing
  systems}}. \bibinfo{pages}{1867--1875}.
\newblock


\bibitem[\protect\citeauthoryear{Quadrianto, Petterson, Caetano, Smola, and
  Vishwanathan}{Quadrianto et~al\mbox{.}}{2010}]%
        {quadrianto2010multitask}
\bibfield{author}{\bibinfo{person}{N. Quadrianto}, \bibinfo{person}{J.
  Petterson}, \bibinfo{person}{T. Caetano}, \bibinfo{person}{A. Smola}, {and}
  \bibinfo{person}{S. Vishwanathan}.} \bibinfo{year}{2010}\natexlab{}.
\newblock \showarticletitle{Multitask learning without label correspondences}.
  In \bibinfo{booktitle}{\emph{Advances in Neural Information Processing
  Systems}}. \bibinfo{pages}{1957--1965}.
\newblock


\bibitem[\protect\citeauthoryear{Schafer and Graham}{Schafer and
  Graham}{2002}]%
        {schafer2002missing}
\bibfield{author}{\bibinfo{person}{J. Schafer} {and} \bibinfo{person}{J.
  Graham}.} \bibinfo{year}{2002}\natexlab{}.
\newblock \showarticletitle{Missing data: our view of the state of the art.}
\newblock \bibinfo{journal}{\emph{Psychological methods}} \bibinfo{volume}{7},
  \bibinfo{number}{2} (\bibinfo{year}{2002}), \bibinfo{pages}{147}.
\newblock


\bibitem[\protect\citeauthoryear{Thrun and Pratt}{Thrun and Pratt}{2012}]%
        {thrun2012learning}
\bibfield{author}{\bibinfo{person}{S. Thrun} {and} \bibinfo{person}{L. Pratt}.}
  \bibinfo{year}{2012}\natexlab{}.
\newblock \bibinfo{booktitle}{\emph{Learning to learn}}.
\newblock \bibinfo{publisher}{Springer Science \& Business Media}.
\newblock


\bibitem[\protect\citeauthoryear{Widmer and R{\"a}tsch}{Widmer and
  R{\"a}tsch}{2012}]%
        {widmer2012multitask}
\bibfield{author}{\bibinfo{person}{C. Widmer} {and} \bibinfo{person}{G.
  R{\"a}tsch}.} \bibinfo{year}{2012}\natexlab{}.
\newblock \showarticletitle{Multitask Learning in Computational Biology.}. In
  \bibinfo{booktitle}{\emph{ICML Unsupervised and Transfer Learning}}.
  \bibinfo{pages}{207--216}.
\newblock


\bibitem[\protect\citeauthoryear{Xu, Zhou, and Tan}{Xu et~al\mbox{.}}{2015}]%
        {xu2015formula}
\bibfield{author}{\bibinfo{person}{J. Xu}, \bibinfo{person}{J. Zhou}, {and}
  \bibinfo{person}{P.-N. Tan}.} \bibinfo{year}{2015}\natexlab{}.
\newblock \showarticletitle{Formula: {F}act{OR}ized {MU}lti-task {L}e{A}rning
  for task discovery in personalized medical models}. In
  \bibinfo{booktitle}{\emph{Proceedings of the 2015 SIAM International
  Conference on Data Mining}}. SIAM, \bibinfo{pages}{496--504}.
\newblock


\bibitem[\protect\citeauthoryear{Zhou, Chen, and Ye}{Zhou
  et~al\mbox{.}}{2011a}]%
        {zhou2011malsar}
\bibfield{author}{\bibinfo{person}{J. Zhou}, \bibinfo{person}{J. Chen}, {and}
  \bibinfo{person}{J. Ye}.} \bibinfo{year}{2011}\natexlab{a}.
\newblock \showarticletitle{Malsar: Multi-task learning via structural
  regularization}.
\newblock \bibinfo{journal}{\emph{Arizona State University}}
  (\bibinfo{year}{2011}).
\newblock


\bibitem[\protect\citeauthoryear{Zhou, Yuan, Liu, and Ye}{Zhou
  et~al\mbox{.}}{2011b}]%
        {zhou2011multi}
\bibfield{author}{\bibinfo{person}{J. Zhou}, \bibinfo{person}{L. Yuan},
  \bibinfo{person}{J. Liu}, {and} \bibinfo{person}{J. Ye}.}
  \bibinfo{year}{2011}\natexlab{b}.
\newblock \showarticletitle{A multi-task learning formulation for predicting
  disease progression}. In \bibinfo{booktitle}{\emph{Proceedings of the 17th
  ACM SIGKDD international conference on Knowledge discovery and data mining}}.
  ACM, \bibinfo{pages}{814--822}.
\newblock


\end{thebibliography}

\end{document}